\documentclass[11pt]{article}

\usepackage[final]{acl}

\usepackage{times}
\usepackage{latexsym}
\usepackage{amsmath}
\usepackage[T1]{fontenc}

\usepackage[utf8]{inputenc}

\usepackage{microtype}

\usepackage{inconsolata}

\usepackage{graphicx}
\usepackage{url}
\usepackage{booktabs}
\usepackage{multirow}
\usepackage{float}
\usepackage{tikz}
\usetikzlibrary{shapes.geometric, arrows.meta, positioning, calc}
\usepackage{listings}
\title{The Coverage Illusion: From Pre-retrieval Routing Failure to Post-retrieval Cascades in a Production RAG System}

\author{Zafar Hussain \\
  Aarhus University, Denmark  \\
  \texttt{zafar@cas.au.dk} \\\And
  Kristoffer Nielbo\\
  Aarhus University, Denmark \\
  \texttt{kln@cas.au.dk} \\} 

\begin{document}
\maketitle
\begin{abstract}
In modern RAG pipelines, query augmentation methods such as HyDE and query expansion are applied to every query, resulting in substantial LLM inference costs and increased end-to-end latency. The empirical justification for this overhead in real production traffic remains largely unexplored. We present a case study of the Danish National Encyclopedia, evaluating five retrieval workflows over 20,000 query–workflow pairs from production traffic and synthetic conditions. In this system, synthetic queries suggest that LLM augmentation is needed for over 90\% of queries to achieve high retrieval coverage. However, under our production deferral policy, only 27.8\% of real user queries need LLM augmentation. We call this gap the \textit{Coverage Illusion} and attribute it to a structural mismatch between synthetic and real query distributions. Pre-retrieval routing cannot resolve this gap, as the need for LLM augmentation is only revealed after searching the index, a result confirmed by our evaluation of four machine learning paradigms. The coverage gap, undetectable from the query alone, motivates a post-retrieval cascade that runs workflows in cheapest-first order and escalates to LLM augmentation only when a step returns no documents. Operating entirely without training overhead or secondary serving infrastructure, the cascade improves quality by +0.140 Composite Overall points over Always-HyDE, reduces latency by 31.8\%, and serves 72.2\% of real user queries without LLM augmentation.
\end{abstract}

\section{Introduction}

Modern Retrieval-Augmented Generation (RAG) systems increasingly use Large Language Models (LLMs) not just for generation but within the retrieval process itself. Techniques such as HyDE \cite{gao-hyde2022} and query expansion \cite{qe-llm2023, qe-rag2023} rewrite or enrich queries before the index is searched, often yielding substantial gains in retrieval coverage but at a high cost in compute and latency. Despite this overhead, both are applied uniformly to every query in production systems \cite{shen-2024-retrieval, adaptive-rag2024}, with no per-query assessment of whether augmentation is necessary.

We study the Danish National Encyclopedia, a production RAG system serving real users across a corpus of \textasciitilde 240,000 curated articles. Two features of this deployment make the question of uniform LLM augmentation consequential. The system defers rather than speculates when retrieval returns no relevant documents, so retrieval \textit{coverage} is the dominant quality factor. Furthermore, real users submit short keyword phrases, quite unlike the verbose conversational queries that dominate English RAG benchmarks \cite{Fan-bench-2024}. Both features motivate the question of whether LLM augmentation can be applied selectively rather than uniformly. A natural solution is pre-retrieval query routing, in which a lightweight classifier assigns each query to the most likely successful workflow \cite{query-routing2025, ragrouter2025}. However, we demonstrate that pre-retrieval routing fails because the need for LLM augmentation depends on the retrieval index rather than the query itself. Our analysis yields three main findings.

\begin{enumerate}
\item \textbf{The Coverage Illusion.} Our synthetic queries indicate LLM augmentation is necessary for over 90\% of queries. On real user queries from the same deployment, only 27.8\% require LLM augmentation under our deferral policy, as retrieval-only Hybrid already returns sources for the remaining 72.2\%.

\item \textbf{Pre-retrieval Routing Fails.} Across four Machine Learning (ML) paradigms (rule-based heuristics, classification, neural fine-tuning, and regression), every approach fails on real user queries. In an entity-heavy reference corpus, whether a query needs augmentation is a property of the index, not of the query text.

\item \textbf{Post-retrieval Cascade.} Since the need for augmentation is revealed only after retrieval runs, the natural response is a cascade in which retrieval proceeds in cheapest-first order and escalates only when a step returns no documents. The post-retrieval cascade improves quality by +0.140 Composite Overall points over the best static baseline, reduces latency by 31.8\%, and invokes LLM augmentation for only 27.8\% of queries.
\end{enumerate}

\section{Related Work}

\paragraph{Synthetic vs. Real Query Evaluation.}
Synthetic queries have become the standard benchmarking tool for RAG systems \cite{synth-rag2025, synth-benchIEEE2025, synth-data-eval2025, eval-rag2025}. This benchmarking practice introduces a fundamental structural mismatch. Synthetic queries tend to be contextually verbose and grammatically complete, whereas real user interactions are characterized by brevity and extreme vocabulary sparsity \cite{eval-rag-synthdata2025, bias-synth-data2025, syntriever2025}. How far this gap skews system-level conclusions, particularly about when retrieval augmentation is necessary, remains unanswered \cite{rag-bench2021}.

\paragraph{Production RAG Systems and Case Studies.}
Despite increasing research on RAG evaluation and adaptive routing \cite{ragsurvey2025, eval-rag2025}, empirical analyses grounded in production settings remain relatively rare. Most studies rely on benchmark datasets that may not reflect real-world query distributions \cite{rag-pipe-2024, rag-pipe-2026}. Our work addresses this gap by analyzing a production system in which query distributions, retrieval characteristics, and quality incentives are drawn from live traffic, enabling us to measure how far synthetic evaluation diverges from real deployment behavior.

\paragraph{Query Routing and Adaptive RAG.}
Adaptive RAG systems are designed to steer each query toward the most appropriate retrieval path. Frameworks like LangChain \cite{langchain2023} and LlamaIndex \cite{llamaindex2023} use hand-crafted heuristics; Gorilla \cite{gorilla2024} and RouteLLM \cite{routellm2024} learn routing policies from data; Self-RAG \cite{selfrag2023} decides dynamically whether to retrieve at all. All these systems commit to a routing decision \textit{before} retrieval, based solely on the input query \cite{routing2025}. To our knowledge, no prior work has questioned whether the query alone provides sufficient signal for this, nor explored routing conditioned on what retrieval returns \cite{ragsurvey2025}.

\paragraph{Advanced Retrieval Strategies for RAG.}
HyDE \cite{gao-hyde2022} and LLM-based query expansion \cite{bertrerank2020, crossencoder2021, qe-rag2023, qe-llm2023} improve retrieval for underspecified queries by rewriting or enriching them. Both methods add substantial inference overhead \cite{retrievalaugmented2020, rag-eval2024}, yet most prior work applies LLM augmentation uniformly to all queries, treating it as universally beneficial \cite{rag-bench2023}.

\section{Experimental Setup}

Our testbed is the Danish National Encyclopedia (lex.dk), a corpus of approximately 240,000 professionally curated articles, semantically chunked, embedded with \texttt{multilingual-e5-large} \cite{Wang2024MultilingualET}, and indexed in a vector database. Three properties suit this deployment for the study. It carries real production traffic, its entity-heavy content produces clear retrieval outcomes, and its stable curation ensures performance differences are attributable to the retrieval workflow rather than corpus noise. Five retrieval workflows operate on this index under four query conditions of 1,000 queries each; each pair is scored by an automated LLM judge, yielding 20,000 evaluated pairs (Figure~\ref{fig:architecture}, Appendix). From these, we construct an oracle dataset that identifies the best-performing workflow for each query, which serves as the benchmark for pre-retrieval routing and the cascade.

\subsection{Query Conditions}

We evaluated four query conditions, each with 1,000 queries. One condition draws from production logs (Real User), while the remaining three are synthetic variants spanning a range of formulation styles, from raw LLM output to natural conversational language (Synth-Polluted, Synth-Keywords, and Synth-Conversational). Full details of each condition, including sources and query statistics, are provided in Appendix~\ref{sec:appendix_datasets} (Tables~\ref{tab:datasets}--\ref{tab:query_stats_full}).

\textbf{Real User} queries come from production search logs and are the primary reference condition for all deployment-relevant conclusions. They have a median length of 5.5 words, typically consist of a bare topic fragment or keyword phrase, as in \textit{Menneskets anatomi} (`Human anatomy'). Users name a topic rather than describe it, a pattern that maps naturally to BM25 \cite{bm25-2009} lexical matching but gives dense retrieval little semantic signal to work with. Their brevity and lexical sparsity make them the most challenging condition for dense retrieval and the only representative of live production traffic. A further 5,000 queries drawn from the same production logs are held out from all routing experiments and used exclusively for the generalization tests in pre-retrieval routing.

\textbf{Synth-Polluted} queries are the raw output of GPT-5.2 \cite{openai-gpt5}, generated from article content using a structured prompt. The outputs are instruction-heavy and chatty, as in \textit{Svar ja eller nej og begrund kort: adskiller impressionismen sig fundamentalt fra tidligere europæiske malestile?} (`Answer yes/no and briefly justify: does impressionism differ fundamentally from earlier European painting styles?'). The remaining two synthetic conditions rewrite these queries (see Appendix~\ref{sec:appendix}).

\textbf{Synth-Keywords} queries are created by rewriting each Synth-Polluted query into a short keyword phrase using GPT-3.5 Turbo~\cite{openai-gpt3}, as in \textit{impressionisme fransk maleri 1800-tal stil} (`impressionism French painting 19th-century style'). They are shorter than Synth-Polluted queries, but more vocabulary-rich than real user queries.

\textbf{Synth-Conversational} queries are created by rewriting each Synth-Polluted query into a natural-language question using GPT-3.5 Turbo~\cite{openai-gpt3}, as in \textit{Hvad adskiller impressionismen fra tidligere europæiske maleritraditioner?} (`What distinguishes impressionism from earlier European painting traditions?'). They closely resemble the query style used in most RAG benchmarks and represent an optimistic setting for LLM-augmented retrieval, as they are rich, contextually complete, and well aligned with the corpus semantics.

\subsection{Retrieval Workflows}

The five workflows span the full cost spectrum and are organized into two tiers (illustrated in Figure~\ref{fig:workflows}, Appendix). Tier~1 retrieves directly from the vector index; Tier~2 first transforms the query by invoking an LLM, and then retrieves.

\textbf{Tier 1 workflows} retrieve directly from the vector index. \textbf{Semantic} embeds the raw query and retrieves the top-$k$ (k = 10) nearest neighbors from the vector index. \textbf{Semantic-CE} does the same but adds a Danish cross-encoder reranker (MiniLM-L6-danish-reranker-v2 \cite{KennethTM}) to rescore the top-$k$ candidates. \textbf{Hybrid} combines BM25 full-text search with dense vector retrieval, merging the top-$k$ ranked lists from both sources via Reciprocal Rank Fusion (RRF) \cite{Cormack2009ReciprocalRF}.

\textbf{Tier 2 workflows} transform the query with an LLM before retrieval. \textbf{QE-CE} prompts an LLM to expand the query with related terms, then retrieves with a hybrid search and reranks with the cross-encoder reranker \cite{KennethTM}. \textbf{HyDE} prompts an LLM to generate a hypothetical answer document, which is then used for retrieval via its embedding.

\subsection{Evaluation Protocol}
\label{sec:eval_protocol}

\textbf{Scale and Parameters.} The full evaluation covers 20,000 query--workflow pairs, comprising five retrieval workflows evaluated across four query conditions of 1,000 queries each. Responses are generated by \texttt{gemma-4-26B-A4B-it}~\cite{google-gemma4} with default parameters; quality scores come from \texttt{Qwen3-32B}~\cite{qwen3-2025} acting as an automated LLM judge, running greedy decoding (\texttt{temperature=0.0}, \texttt{max\_tokens=4096}) for deterministic, reproducible evaluations. All reported latency figures are end-to-end wall time, including LLM generation. To check that our findings do not depend on the specific judge, we replicated the evaluation with \texttt{Qwen3.6-27B}~\cite{qwen3-2025} (comparison of both given in Appendix~\ref{sec:appendix_qwen36}).

\textbf{Metrics.} Each response receives three scores from the automated judge on a 1--5 scale, covering faithfulness (claims grounded in retrieved sources), answer relevance (response addresses the question), and retrieval quality (documents are on-topic). Our primary metric, \texttt{composite\_overall} (\textbf{CO}), averages faithfulness and answer relevance across all queries; retrieval quality is reported separately as a process diagnostic. Queries returning no documents are assigned CO = 1.0, the minimum, since the system must defer rather than generate an ungrounded response. We also report \texttt{composite\_when\_answered} (\textbf{CWA}), computed only over answered queries, which isolates per-answer quality from the coverage effect. The binary \texttt{has\_sources} flag (1 for answered queries and 0 for deferred ones) partitions these two groups and serves as the escalation trigger in the cascade.

\section{The Coverage Illusion}

We establish per-workflow baselines on all 1,000 real user queries. Comparison with synthetic conditions then reveals how far standard benchmarks overstate the need for LLM augmentation.  

\subsection{Static Baselines}

Table~\ref{tab:baselines} reports results when each workflow is applied uniformly to all 1,000 real user queries. CO follows coverage closely across all five systems, from Semantic (62.6\%, CO=3.058) up to HyDE (86.4\%, CO=3.944), confirming that coverage is the primary driver of composite quality. The narrow CWA spread (4.287--4.407, a range of just 0.12 points) shows that once retrieval succeeds, all five workflows produce comparably good answers. Always-HyDE incurs 96.2s of end-to-end latency, 2.6$\times$ more than the Hybrid baseline (37s), without a commensurate gain in answer quality. The oracle ceiling (CO=4.404) marks the best achievable quality under perfect per-query routing.

\begin{table}[h]
\centering
\scriptsize
\setlength{\tabcolsep}{4pt}
\resizebox{\columnwidth}{!}{%
\begin{tabular}{lp{2.85cm}rrrr}
\toprule
\textbf{Workflow} & \textbf{Method} & \textbf{CO} & \textbf{CWA} & \textbf{Cov.} & \textbf{Lat.} \\
\midrule
\multicolumn{6}{l}{\textit{Tier 1: Retrieval-Only}} \\
\midrule
Semantic    & Dense vector search        & 3.058 & 4.287 & 62.6\% & 33s \\
Semantic-CE & Dense + CE rerank       & 3.086 & 4.321 & 62.8\% & 33s \\
Hybrid      & BM25 + Dense + RRF        & 3.408 & 4.335 & 72.2\% & 37s \\
\midrule
\multicolumn{6}{l}{\textit{Tier 2: LLM-Augmented}} \\
\midrule
QE-CE       & LLM expansion + CE rerank  & 3.601 & 4.338 & 77.9\% & 60s \\
HyDE        & Hypothetical doc.\ embedding    & 3.944 & 4.407 & 86.4\% & 96s \\
\midrule
Oracle      & ---                          & 4.404 & 4.940 & ---      & ---   \\
\bottomrule
\end{tabular}
}
\caption{Retrieval workflows and static performance; latency is end-to-end wall time, including generation.}
\label{tab:baselines}
\end{table}

\subsection{Coverage Drives Quality}

LLM-augmented workflows score higher because they retrieve sources more frequently, not because they generate better answers. The binary \texttt{has\_sources} flag explains 69--83\% of overall quality variance (mean R\textsuperscript{2}=0.77; Appendix~\ref{sec:appendix_variance}). To quantify this, we decompose the difference in composite overall score ($\Delta\text{CO}$) between HyDE and the Semantic baseline using a first-order approximation. The decomposition is defined as:
$$\Delta\text{CO} \approx \Delta\text{cov} \times \bar{Q}_{\text{baseline}} + \text{cov}_{\text{baseline}} \times \Delta\bar{Q}_{\text{ans}} + \epsilon$$

where $\Delta\text{cov}$ is the coverage gain, $\bar{Q}_{\text{baseline}}$ is the baseline's mean per-answer quality, $\text{cov}_{\text{baseline}}$ is the baseline's coverage rate, $\Delta\bar{Q}_{\text{ans}}$ is the improvement in per-answer quality, and $\epsilon$ is a residual term that captures the marginal performance of queries that are resolved only by the augmented workflow. Under this formalization, coverage expansion contributes $+1.020$ CO points and improved synthesis contributes $+0.076$ CO points. The sum of these two components exceeds the observed total because the residual term $\epsilon = -0.209$ is negative. This occurs because the queries HyDE reaches that Semantic does not tend to be harder, and their actual per-answer quality falls below Semantic's average, causing the approximation to overshoot. HyDE’s advantage is almost entirely due to retrieving sources that Semantic misses, with minimal impact from synthesis quality (Appendix~\ref{sec:appendix_decomp}).


\subsection{The Gap Between Synthetic and Real}

\begin{figure}[h]
    \centering
    \includegraphics[width=0.97\columnwidth]{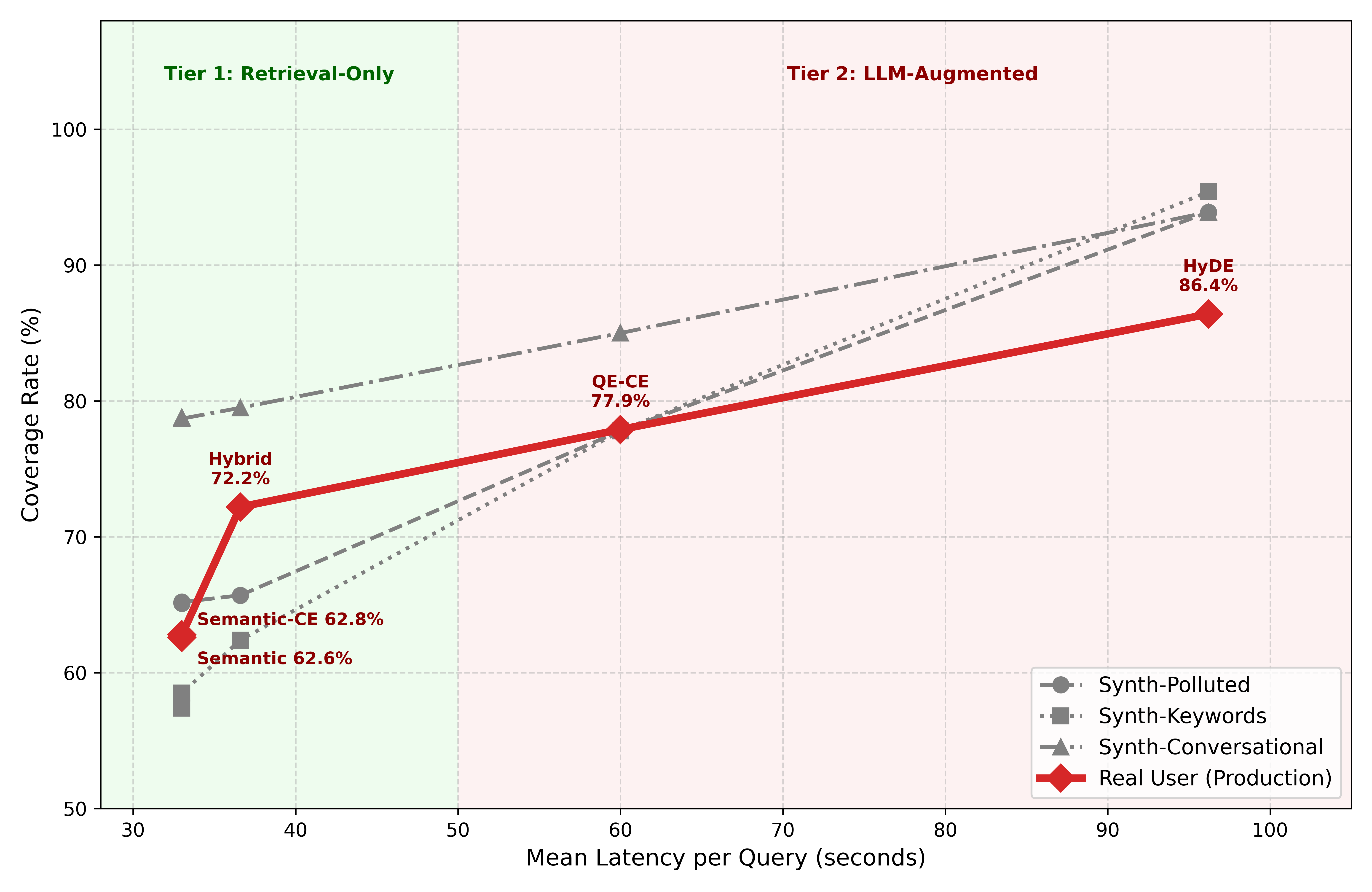}
    \caption{The Coverage Illusion, showing retrieval coverage by workflow and query condition.}
    \label{fig:coverage_illusion}
\end{figure}

On synthetic queries, LLM-augmented workflows achieve 93--95\% coverage, far ahead of retrieval-only alternatives; on real user queries, this advantage collapses. Retrieval-only Hybrid already attains 72.2\% coverage, and HyDE raises it only to 86.4\% (a 14.2 point gain), versus a 32.9 point HyDE--Hybrid gap on Synth-Keywords. This disparity is the Coverage Illusion in practice (Figure~\ref{fig:coverage_illusion}). In our evaluation, the synthetic query conditions imply augmentation is necessary for over 90\% of queries, yet on real user queries from the same deployment, only 27.8\% require escalation under our deferral policy. The underlying cause is a structural difference between query types. Real user queries have a mean length of 7 words, typically a keyword phrase that BM25 maps directly to the relevant article. Synthetic queries average 10--20 words and use broader, rich vocabulary, which reduces lexical overlap with the corpus and makes retrieval-only approaches appear less effective than they are on real traffic.

\section{The Limits of Pre-retrieval Routing}
\label{sec:routing}

If 72.2\% of real user queries do not require LLM augmentation, a lightweight pre-retrieval classifier could, in principle, identify those queries in advance and skip the LLM augmentation step entirely. We tested this across four machine learning paradigms of increasing complexity. To focus on queries where the workflow choice actually matters, we evaluated on a subset of 2,206 oracle queries where the best and worst workflow diverge by at least 1.5 CO points; among these, 66.4\% have an LLM-augmented oracle label and 33.6\% a retrieval-only label (Table~\ref{tab:oracle_label_dist}). For real user queries specifically, 609 of 1,000 meet this threshold, split nearly evenly at 52.4\% LLM-augmented and 47.6\% retrieval-only. We held out 20\% of the dataset as a stratified test set for all ML experiments, giving $n=442$ for classification and regression models; the rule-based heuristics use a slightly larger split ($n=494$) because oracle labels are recomputed with a looser tie-breaking threshold.

Since the high-contrast subset filters out queries where workflows tie, Always-HyDE on this subset scores CO=4.258, above the full-set figure of 3.944 in Table~\ref{tab:baselines}; queries where HyDE offers no advantage over cheaper workflows are excluded. Against this elevated baseline, every routing model fell well short of the oracle ceiling, and all models degraded further when evaluated on real user queries.

\subsection{Rule-Based Heuristic Routing}

Handcrafted string rules (query length, question marks, Danish question words) reach \textbf{38.4\% accuracy}, just above the uniform-random baseline of 20\% for five classes. The heuristic routes queries to three of the five workflow classes; QE-CE and Semantic-CE receive zero correct predictions. Per-class recall is poor across the covered workflows (Semantic: 63.2\%, HyDE: 37.3\%, Hybrid: 25.8\%). On the stratified test split ($n=494$), the router achieves CO=4.202, \textbf{below} Always-HyDE (4.258).

\subsection{Classification Routing Models}

Table~\ref{tab:ml_routers} shows that only the simplest feature configurations outperform Always-HyDE (CO=4.258). Logistic Regression (LR) with TF-IDF character n-grams achieves the best result (CO=4.269, $\Delta=+0.011$), while LR with surface features (character counts, question marks, stop-word ratios) follows closely (CO=4.260, $\Delta=+0.002$). LR with word n-grams falls below the baseline (CO=4.240, $\Delta=-0.018$). Even the best model recovers just 2.4\% of the 0.475-point gap between Always-HyDE and the oracle ceiling.

\begin{table}[h]
\centering
\small
\begin{tabular}{lrr}
\toprule
\textbf{Model} & \textbf{CO} & \textbf{$\Delta$ vs HyDE} \\
\midrule
LR / TF-IDF char (3--5)grams & 4.269 & +0.011 \\
LR / surface features         & 4.260 & +0.002 \\
LR / TF-IDF word (1--2)grams  & 4.240 & $-$0.018 \\
GBM / TF-IDF + surface        & 4.177 & $-$0.081 \\
GBM / multilingual embed.      & 4.184 & $-$0.074 \\
LR / multilingual embed.      & 3.850 & $-$0.408 \\
\midrule
Always-HyDE (baseline)        & 4.258 & ---      \\
Oracle (ceiling)              & 4.733 & +0.475 \\
\bottomrule
\end{tabular}
\caption{Classification model router performance on oracle test split ($n=442$, high-contrast queries only).}
\label{tab:ml_routers}
\end{table}

Richer features and greater model capacity consistently worsen routing quality. A Gradient Boosting Machine (GBM) with TF-IDF and surface features scores CO=4.177 ($\Delta=-0.081$), below every LR variant. A GBM over multilingual embeddings reaches only CO=4.184 ($\Delta=-0.074$). LR over multilingual embeddings (paraphrase-multilingual-MiniLM-L12-v2) \cite{reimers-gurevych-2019-sentence} is the worst configuration (CO=3.850, $\Delta=-0.408$); minority classes such as QE-CE (21\% recall) and Semantic-CE (29\% recall) are predicted nearly at random. Across all six models, richer representations yield less routing signal than character n-grams because multilingual embeddings are designed to suppress surface differences in favor of semantic similarity, removing the surface variation that weakly tracks whether a query will match the index by keyword.

On 5,000 unseen real user queries, the best LR model (TF-IDF character n-grams) routes over 90\% of queries to HyDE, capturing under 5\% of the augmentation-free opportunity versus the 33.6\% oracle ceiling (Table~\ref{tab:oracle_label_dist}). Even for one-to-three-word queries, where BM25 coverage is highest, the most are still routed to HyDE, confirming that query length is not a reliable routing signal. Retraining on the 609 real user queries reduces domain shift, but augmentation-free routing reaches only \textbf{13.4\%} (of the oracle ceiling), limited to the two cheapest workflows. The cascade (Section~\ref{sec:cascade}) achieves 72.2\% augmentation-free routing with no training, 5.4$\times$ rate of the best ML router.

\subsection{Neural Encoder Fine-Tuning}

We fine-tuned all-MiniLM-L6-v2 \cite{reimers-gurevych-2019-sentence} for 5 epochs on the oracle labels. Evaluated on 5,000 blind real user queries, the model routed 73.8\% of queries to HyDE and 26.2\% to Hybrid, assigning none to the other workflows. Training loss fell by 16\% (1.224$\to$1.025) over 5 epochs, pointing to signal failure rather than insufficient training. Weighting CO by routing share yields an estimated CO of \textbf{3.804}, below Always-HyDE. The model separates broad augmentation tiers but cannot discriminate within a tier.

\subsection{Regression-Based Routing Models}
We trained a regression model to predict the quality score for each workflow based on the query text. The system defaults to HyDE; a cheaper alternative replaces it only when HyDE's predicted score exceeds the Alternative's by at least $\delta$ points, a threshold that guards against low-confidence substitutions. Table~\ref{tab:regression_pareto} (Appendix~\ref{sec:appendix_tables}) reports the Pareto frontier for Ridge regression under 5-fold cross-validation.

The best operating point ($\delta=0.75$) captures only \textbf{3.7\%} of the oracle gap, redirecting just 4.1\% of queries away from HyDE, well short of the 33.6\% retrieval-only ceiling (Table~\ref{tab:oracle_label_dist}; Appendix~\ref{sec:appendix_oracle_labels}). GBM regression with far greater capacity performs worse (CO=4.222, $\delta=2.00$, capturing 1.4\% of the oracle gap), confirming that the bottleneck is the missing signal, not model complexity. Above $\delta=0.75$, the router collapses to Always-HyDE. Below $\delta=0.50$, quality falls below the HyDE baseline, as the low threshold causes the model to route too many queries away from HyDE, including queries that genuinely needed it.

\subsection{Root Cause: Query Content Limitations}
All four paradigms fail for the same reason. The need for LLM augmentation is determined by what the retrieval index contains, not by what the query says. Whether a given query needs augmentation depends on whether the index holds directly relevant content for it, which is unknowable before retrieval runs. No surface feature, embedding, or learned classifier can substitute for this post-retrieval signal. We hypothesize that this limitation generalizes to any entity-heavy reference corpus where real users submit direct factual lookups.

\section{Post-Retrieval Cascade Architecture}
\label{sec:cascade}

\subsection{Design}
\begin{figure}[h]
  \centering
  \includegraphics[width=0.99\columnwidth]{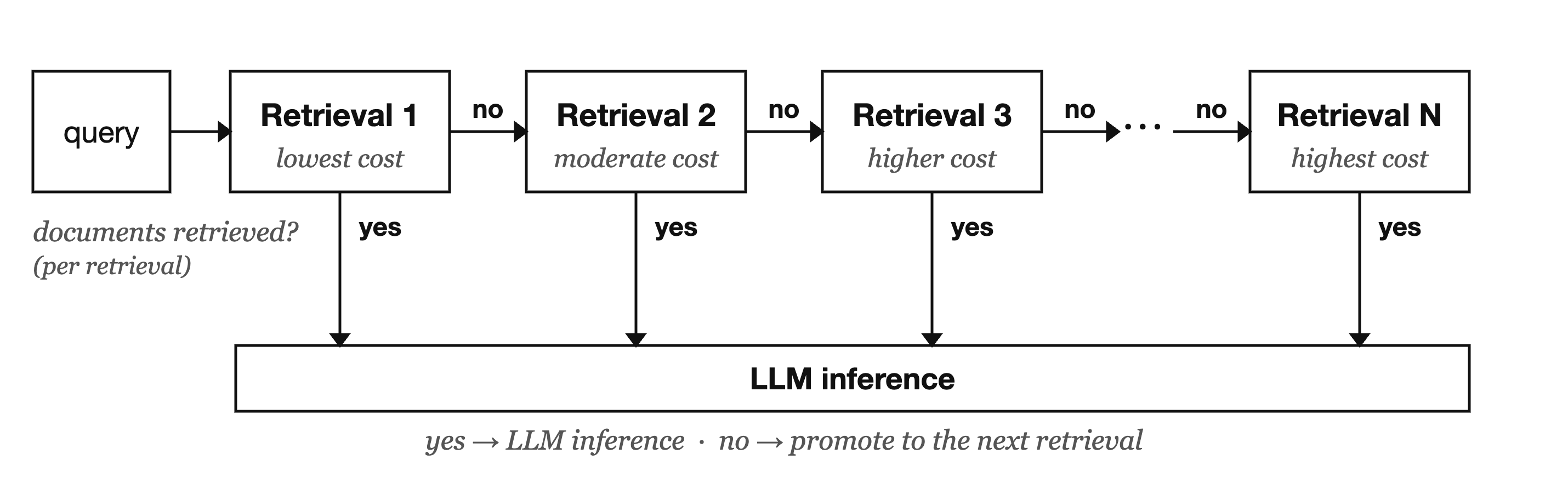}
  \caption{\small Post-retrieval cascade execution flow.}
  \label{fig:cascade}
\end{figure}

The failure of pre-retrieval routing demonstrates that query-side features cannot anticipate the contents of the retrieval index, making pre-retrieval prediction inherently limited. The cascade instead responds to the actual retrieval outcome, running workflows in order of increasing cost and escalating only when a step returns no documents. Its value is empirical rather than architectural, quantifying the margin by which a reactive policy outperforms any trained routing alternative. As shown in Figure~\ref{fig:cascade}, if retrieval succeeds the LLM generates an answer and the process stops; otherwise, the cascade advances to more expensive workflows, ultimately deferring if no evidence is found.

Formally, let $q$ be the incoming query and $\mathcal{W} = (w_1, w_2, \dots, w_n)$ the ordered sequence of retrieval workflows, sorted by ascending cost. For each $w_i \in \mathcal{W}$, let $R(q, w_i)$ denote the retrieved document set. 

The cascade evaluates a binary success indicator $I(q, w_i)$:

$$I(q, w_i) = \begin{cases} 1 & \text{if } |R(q, w_i)| > 0 \\ 0 & \text{otherwise} \end{cases}$$

If $I(q, w_i) = 1$, the cascade terminates, and the retrieved document set $R(q, w_i)$ is passed to the generation model. If $I(q, w_i) = 0$, the system escalates, initiating retrieval for the next workflow $w_{i+1}$ in the sequence. If all workflows return empty results, the system applies the deferral policy.

\subsection{Efficiency of the Stopping Condition}

The \texttt{has\_sources} check is an $O(1)$ array operation on the retriever's return value, as no LLM is invoked at this stage. The binary threshold is also empirically optimal. To confirm this, we ran offline simulations with relaxed escalation thresholds. Escalating when fewer than two or three documents were retrieved, rather than when none were found, reduced composite quality from 4.03 to 3.98. When at least one well-matched document is already available, a hypothetical document adds noise rather than coverage.

\section{Results}

\subsection{Cascade Performance: Real User Queries}

Across all 1,000 real user queries, the cascade improves quality, cuts latency, and limits LLM augmentation to 27.8\% of queries, compared to 100\% under Always-HyDE (Table~\ref{tab:cascade_results}). Composite score rises from CO=3.944 to CO=4.084 ($\Delta=+0.140$), end-to-end latency falls 31.8\% from 96.2s to 65.6s, and the language model is invoked only for queries where Hybrid retrieval fails. These gains do not trade quality for speed. 72.2\% of real user queries stop at the first cascade step and never incur the LLM transformation that Always-HyDE applies to all queries. The lower Always-HyDE baseline of CO=3.944 here, compared with CO=4.258 in Section~\ref{sec:routing}, reflects the inclusion of the full query set, which restores tied queries for which HyDE offers no advantage over cheaper workflows.

\begin{table}[h]
\centering
\scriptsize
\begin{tabular}{lrrrrr}
\toprule
\textbf{Strategy} & \textbf{CO} & \textbf{$\Delta$CO} & \textbf{Latency} & \textbf{Saved} & \textbf{Aug.\%} \\
\midrule
Always-HyDE        & 3.944 & ---      & 96.2s & ---     & 100\% \\
\textbf{Cascade}   & \textbf{4.084} & \textbf{+0.140} & \textbf{65.6s} & \textbf{31.8\%} & \textbf{27.8\%} \\
Oracle ceiling     & 4.404 & +0.460 & ---     & ---     & ---     \\
\bottomrule
\end{tabular}
\caption{Best cascade vs.\ Always-HyDE on all 1,000 real user queries.}
\label{tab:cascade_results}
\end{table}

27.8\% of all queries for which Hybrid returns no documents escalate to QE-CE or HyDE. The quality gain over Always-HyDE has two sources. First, the cascade avoids the quality degradation that HyDE causes on queries that keyword matching resolves cleanly. When the relevant article is a direct lexical match, HyDE's hypothetical document redirects retrieval toward adjacent but less relevant sources (the per-query breakdown is in Section~\ref{sec:discussion}). Second, the cascade reserves HyDE's coverage for queries that genuinely need it.

Table~\ref{tab:cascade_variants} compares all tested cascade orderings. The three-step configuration (Hybrid $\to$ QE-CE $\to$ HyDE) achieves the highest quality (CO=4.084). The two-step variant (Hybrid $\to$ HyDE) saves an additional 2.8 seconds at a cost of only 0.016 CO points; inserting Semantic-CE as an intermediate step incurs additional latency without improving quality. The cascade outperforms Always-HyDE in every ordering tested, confirming the gains are robust to the specific steps chosen. The oracle ceiling (CO=4.404) marks the best achievable quality under perfect per-query routing; every pre-retrieval method we tested covered at most 2.2\% of this gap (Table~\ref{tab:routing_vs_cascade}), whereas the cascade, without training or additional infrastructure, covers 30\%.

\begin{table}[h]
\centering
\scriptsize
\setlength{\tabcolsep}{4pt}
\begin{tabular}{p{3.8cm}rrrr}
\toprule
\textbf{Cascade} & \textbf{CO} & \textbf{$\Delta$CO} & \textbf{Lat.} & \textbf{Saved} \\
\midrule
\textbf{Hyb$\to$QE-CE$\to$HyDE}& \textbf{4.084} & \textbf{+0.140} & \textbf{65.6s} & \textbf{31.8\%} \\
Hyb$\to$Sem-CE$\to$QE-CE$\to$HyDE & 4.078 & +0.134 & 68.6s & 28.7\% \\
Hyb$\to$HyDE                       & 4.068 & +0.124 & 62.8s & 34.7\% \\
Sem$\to$Hyb$\to$QE-CE$\to$HyDE & 4.050 & +0.106 & 68.9s & 28.4\% \\
\midrule
Always-HyDE                           & 3.944 & ---      & 96.2s & ---     \\
\bottomrule
\end{tabular}
\caption{Cascade variants on real user queries.}
\label{tab:cascade_variants}
\end{table}

The +0.140 advantage is observed across all 1,000 real user queries without filtering. Restricting to the 609 real user queries that form the high-contrast routing subset (where workflows diverge by at least 1.5 CO points), the advantage grows to $\Delta=+0.241$ (CO=3.936 vs.\ HyDE CO=3.695; Table~\ref{tab:cross_condition_routing}). The routing subset retains only queries where the workflow choice affects quality.

\subsection{Pareto Dominance}

\begin{figure}[h]
    \centering
    \includegraphics[width=0.99\columnwidth]{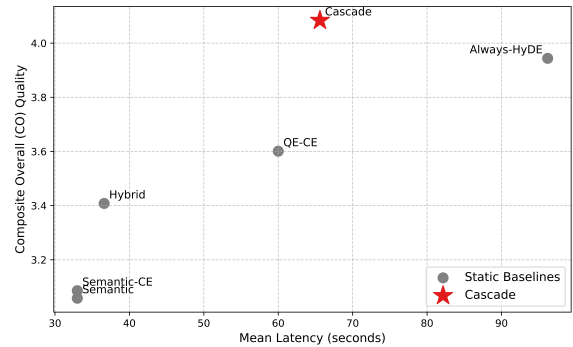}
    \caption{Cost vs.\ Quality on real user queries.}
    \label{fig:pareto}
\end{figure}

Figure~\ref{fig:pareto} shows that the cascade is the only point combining strictly higher quality with strictly lower cost. The 30.6-second saving is real computation avoided, as 72.2\% of queries stop at Hybrid and never invoke LLM augmentation. Among the static baselines, Hybrid offers comparable speed but fails to retrieve sources for 27.8\% of queries; the cascade gives Hybrid's speed to the majority while reserving HyDE for the minority that needs it.

\subsection{Cross-Condition Analysis}

Table~\ref{tab:cross_condition} (Appendix~\ref{sec:appendix_cross_condition}) evaluates the cascade across all four query conditions. On real user queries, the cascade wins by +0.140 CO with 31.8\% latency savings. Even on Synth-Conversational queries, the condition most favorable to LLM augmentation, the cascade matches HyDE (+0.010~CO) while saving 41.9\% in latency. On Synth-Keywords and Synth-Polluted queries, Always-HyDE outscores the cascade by 0.139 and 0.048 CO points respectively, though the cascade still saves roughly 29--30\% in latency in both cases.

This reversal is the Coverage Illusion in action. The condition under which HyDE wins, rich and verbose queries, is precisely what synthetic benchmarks over-represent. Even Synth-Conversational, the most linguistically complete condition, yields only a +0.010 CO cascade advantage, confirming that vocabulary-index alignment, rather than query length or grammatical form, determines whether augmentation is needed.

\subsection{Human Evaluation}
\label{sec:human_eval}

Three independent annotators scored 100 stratified query--response pairs on faithfulness, answer relevance, and retrieval quality using the same 1--5 scale as the automated judge. The automated judge slightly overestimates human scores across all three metrics ($\Delta$ = +0.04 to +0.14 points), with Spearman correlations ranging from $\rho$=0.59 to $\rho$=0.78 (Table~\ref{tab:human_eval_metrics}, Appendix~\ref{sec:appendix_human_eval}). Workflow rankings are consistent between human and automated evaluation, confirming that the cascade's advantage over Always-HyDE is not an artifact of the automated judge. This is corroborated by a second LLM judge, \texttt{Qwen3.6-27B}, which replicates the workflow ranking and coverage-dominance pattern of the primary \texttt{Qwen3-32B} judge (Table~\ref{tab:qwen36_real}, Appendix~\ref{sec:appendix_qwen36}).

\section{Discussion}
\label{sec:discussion}

The cascade's advantage over Always-HyDE is not merely a matter of efficiency. For a significant proportion of queries, the uniform application of HyDE actively degrades response quality. Among the 722 real user queries where Hybrid retrieved sources, the per-query comparison reveals three distinct outcomes. In 59.6\% of cases, the two workflows scored the same, indicating that LLM augmentation added nothing. In 21.2\% of cases, Hybrid outperformed HyDE by a mean of 2.010 CO points. HyDE prevailed in only 19.3\% of cases, with an average margin of 1.324 CO points.

When a direct lexical match successfully locates the target article, the language model's hypothetical document introduces semantic drift, steering retrieval toward adjacent, less relevant passages rather than the precise target source. The result is a quality penalty. HyDE's per-query losses in this regime (mean 2.010 CO points) systematically exceed its gains (mean 1.324 CO points), even though this asymmetry is invisible in aggregate metrics because it affects only the minority of queries where Hybrid already succeeds (Appendix~\ref{sec:appendix_hybrid_vs_hyde}).

These findings challenge the assumption that more inference always improves retrieval. In our encyclopedia deployment, we believe that, as in other entity-heavy reference systems, most queries are simple lookups that keyword matching can handle. Routing every query through LLM augmentation is inefficient, since Hybrid alone retrieves sources for 72.2\% of queries, and in those, HyDE provides no improvement in 59.6\% of cases and actually reduces quality in 21.2\% (Appendix~\ref{sec:appendix_hybrid_vs_hyde}).

\begin{table}[h]
\centering
\scriptsize
\begin{tabular}{lrrr}
\toprule
\textbf{Strategy} & \textbf{Aug.-free\%} & \textbf{CO} & \textbf{Training} \\
\midrule
Always-HyDE (baseline)     &  0.0\% & 3.944 & None \\
ML Router (all conditions) & $<$5.0\% & 3.954 & 1,764 labels \\
ML Router (real user)     & 13.4\% & 3.920 & 609 labels \\
Neural Router (fine-tuned) & 26.2\% & 3.804 & 1,764 labels \\
\textbf{Cascade}    & \textbf{72.2\%} & \textbf{4.084} & \textbf{None} \\
Oracle (ceiling)           & 33.6\% & 4.404 & Perfect labels \\
\bottomrule
\end{tabular}
\caption{Pre-retrieval routing vs.\ cascade}
\label{tab:routing_vs_cascade}
\end{table}

Table~\ref{tab:routing_vs_cascade} makes the contrast concrete. An ML router requires oracle labels, model training, and serving infrastructure, and reaches at most 13.4\% augmentation-free routing. The cascade requires only a Boolean check on the retriever's output, and reaches 72.2\% augmentation-free routing at a higher quality score than any ML alternative. Its augmentation-free rate (72.2\%) exceeds the 33.6\% retrieval-only oracle ceiling because that ceiling is computed on high-contrast queries; beyond that subset, Hybrid still finds relevant sources for many queries where no workflow has a strong advantage.

\section{Conclusion}

We document the \textit{Coverage Illusion} in a production RAG system. Synthetic query conditions suggest that LLM augmentation is necessary for the majority of queries; on real user queries from the same deployment, retrieval-only Hybrid already returns sources for 72.2\%, so only 27.8\% require escalation under our deferral policy.

No pre-retrieval routing method bridges this gap. Across all four ML paradigms tested, the failure is consistent. Whether augmentation is needed for a given query depends on the retrieval index, which becomes apparent once retrieval has run.

The cascade reactively routes queries with an $O(1)$ check on the retriever's output, requiring no training or serving infrastructure. For real user queries, it improves quality by +0.140 CO points and reduces latency by 31.8\% compared to Always-HyDE, using LLM augmentation for just 27.8\% of queries. The design is workflow-agnostic and adaptable to any ordered set with a binary escalation condition. RAG evaluation benchmarks dominated by synthetic queries risk overstating how often LLM augmentation is necessary, particularly for deployments where real users submit short keyword lookups against an entity-rich index. Practitioners should validate the augmentation need on real user traffic before assuming benchmark results carry over to their production setting.

\section{Limitations}

\paragraph{Single Corpus and Deployment Setting.}
All experiments were conducted on a single encyclopedia. The most consequential constraint on generalization is the deferral policy, under which failed retrievals receive CO=1.0 (Section~\ref{sec:eval_protocol}), making coverage the dominant quality factor by construction. Permissive systems that fall back on parametric generation face a smaller coverage penalty, so augmentation would appear more valuable there.

Three further constraints are deployment-specific. Real encyclopedia users submit short keyword phrases at rates that deviate from standard benchmarks (40.4\% keyword-style, 21.0\% well-formed), making hybrid retrieval inherently competitive; conversational or long-form settings may present a different picture. Danish is also morphologically rich, and BM25 lexical matching may differ from English in ways that affect Tier~1 vs.\ Tier~2 performance; we make no claims about other languages.

\paragraph{Cascade Worst-Case Overhead.}
Of the 27.8\% of queries that Hybrid does not resolve, approximately 12.3\% stop at QE-CE, while the remaining 15.5\% escalate through all cascade steps to HyDE. That 15.5\% face a mean cumulative latency of $\sim$161s, because each intermediate step contributes wall time even when it returns no documents. In production, where failed steps skip generation entirely (incurring only retrieval time for Hybrid, and LLM expansion plus retrieval for QE-CE), this worst case drops to approximately 125s ($\sim$1.3$\times$ the 96.2s HyDE baseline).

\paragraph{LLM-as-Judge Bias.}
All primary quality scores were produced using an LLM acting as a judge. While automated judges are widely adopted to improve scalability, they can exhibit systematic preferences for longer outputs and for stylistic alignment. We address this by conducting a human validation study with three independent annotators scoring 100 stratified query--response pairs. Spearman correlations between human and automated scores range from $\rho$=0.59 (faithfulness) to $\rho$=0.78 (answer relevance), with a mean absolute error of 0.17--0.21 points on the 1--5 scale. The LLM judge overestimates by at most 0.14 points. Crucially, workflow rankings are consistent across both evaluation paradigms, confirming that the dominance of the binary coverage signal is an intrinsic system property rather than an evaluation artifact.

\section{Acknowledgments}
We would like to thank Simon Enni (Aarhus University) for reviewing the paper and Yevhen Kostiuk (Aarhus University) for generating the synthetic-polluted dataset. We would like to thank Maja Bressendorff and others (at lex.dk) for providing the support in data and annotations.  

Zafar Hussain and Kristoffer Nielbo are funded by the LEX project (Aage and Johanne Louis-Hansens Foundation (25-1-17733) and Augustinus Foundation (2025-0299)) and the Danish Foundation Models project (4378-00001B). Kristoffer Nielbo is additionally funded by the Danish National Research Foundation (DNRF193) and The Carlsberg Foundation (CF23-1583).

\bibliography{custom}

\appendix

\begin{figure*}[t]
\centering
\includegraphics[width=\textwidth]{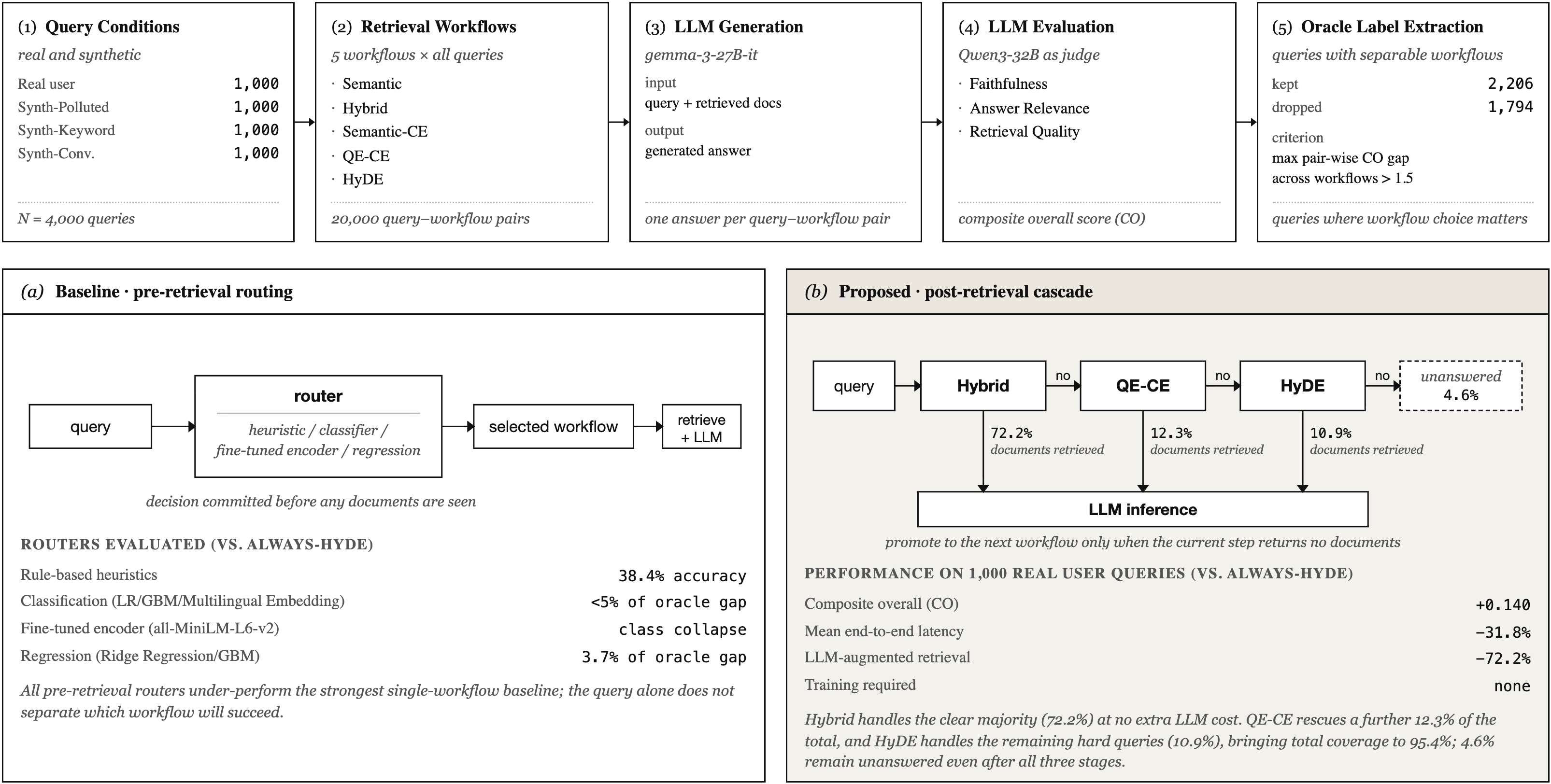}
\caption{Experimental architecture: oracle benchmark construction (top), pre-retrieval routing baselines (a), and post-retrieval cascade (b).}
\label{fig:architecture}
\end{figure*}

\begin{figure*}[t]
\centering
\includegraphics[width=\textwidth]{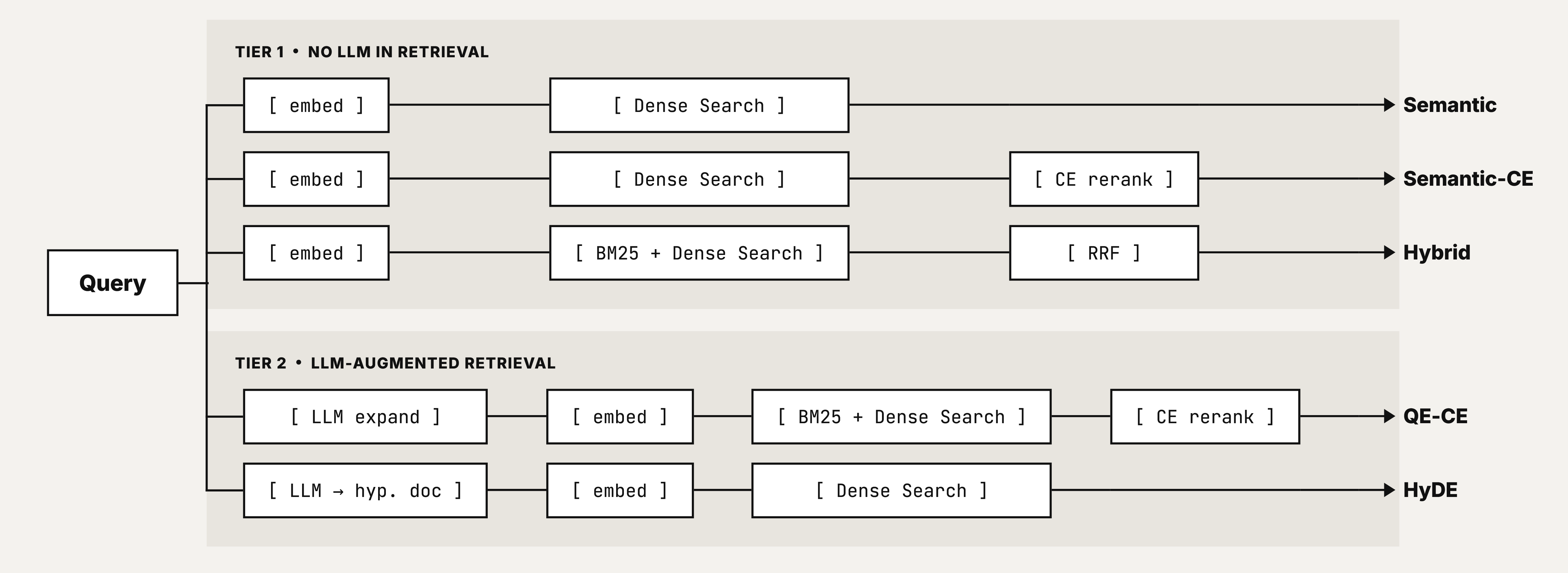}
\caption{Retrieval workflow tiers: Tier~1 (retrieval-only) and Tier~2 (LLM-augmented).}
\label{fig:workflows}
\end{figure*}

\section{Query Generation Prompts}
\label{sec:appendix}

The following prompts produced the three synthetic query conditions; all derive from the same 1,000 article-grounded query--answer pairs and differ only in how the query was reformulated.

\subsection{Synth-Polluted - Generation Prompt for GPT-5.2}
\label{sec:prompt-polluted}

Synth-Polluted queries are the \emph{raw output} of GPT-5.2. Each call received a full article, a randomly sampled query style from a fixed taxonomy, and a set of real production queries as stylistic anchors. The prompt was built around the following five components.

\begin{description}
  \item[Role \& context.] Senior RAG evaluation data scientist at a national encyclopedia; the objective is to build an evaluation dataset that tests retrieval accuracy, multi-section synthesis, and grounded generation.

  \item[Query constraints.] Each query must be (i)~fully self-contained and interpretable in isolation, (ii)~written in Danish, (iii)~grounded strictly in the provided article, (iv)~free of answer hints and article self-references, and (v)~independent of all other queries in the batch.

  \item[Complexity annotation.] Each query was labeled along three dimensions (\textit{information\_composition}, \textit{article\_coverage}, and \textit{reasoning\_load}).

  \item[Structure \& information types.] Two template variables, {structure\_instructions} and {infotype\_instructions}, injected a fixed taxonomy of query structures (e.g., yes/no, list, comparison) and information types (e.g., causal, temporal). The model was instructed \emph{not} to force a type the article did not support.

  \item[Output format.] A JSON list of objects, each containing query, answer, sections\_for\_answer, and a tags block holding info\_types, styles, structures, and the complexity annotation.
\end{description}


\subsection{Synth-Keywords - Keyword Abstraction Prompt for GPT-3.5 Turbo}
\label{sec:prompt-keywords}

Each Synth-Polluted query was rewritten into a keyword query of 5-10 words using the following prompt.

\begin{lstlisting}
You are a query abstractor. The following is a synthetically generated query that contains unnatural instructions or chatty prefixes. Rewrite it into a short keyword-style search phrase (5-10 words). Do NOT use full sentences, questions, or unnecessary verbs. Use only the core nouns and topic terms. Optionally use a colon to separate the main subject from descriptors.
Original Query: {original_query} Keyword Search Phrase (in Danish):
\end{lstlisting}

\subsection{Synth-Conversational - Conversational Rewrite Prompt for GPT-3.5 Turbo}
\label{sec:prompt-conv}

Each Synth-Polluted query was rewritten as a natural, conversational Danish question using the following prompt.

\begin{lstlisting}
You are a query cleaner. The following is a synthetically generated query that contains unnatural instructions or chatty prefixes. Rewrite it into a single, natural Danish QUESTION (like what someone would ask a person). Do NOT summarize it into keywords. Do not use abstract compound nouns. Just frame the core intent as a simple, natural, conversational question. Ensure it ends with a question mark.
Original Query: {original_query} Natural Conversational Question (in Danish):
\end{lstlisting}

\section{Datasets and Query Characterization}
\label{sec:appendix_datasets}

Table~\ref{tab:datasets} summarizes the four query conditions. Table~\ref{tab:query_stats_full} gives full distributional statistics. Real user queries stand out for their brevity. Among them, 8.8\% are single words, 40.4\% are keyword-style, and only 21.0\% are well-formed sentences, a sharp contrast with all three synthetic conditions.

\begin{table}[H]
\centering
\scriptsize
\resizebox{\columnwidth}{!}{%
\begin{tabular}{lrll}
\toprule
\textbf{Condition} & \textbf{n} & \textbf{Source} & \textbf{Style} \\
\midrule
Real User       & 1,000 & Production logs  & Short, keyword-heavy \\
Synth-Polluted   & 1,000 & GPT-5.2          & Instruction-heavy, chatty \\
Synth-Keywords   & 1,000 & GPT-3.5 Turbo    & Topic-focused keywords \\
Synth-Conv.    & 1,000 & GPT-3.5 Turbo    & Natural-language questions \\
\bottomrule
\end{tabular}%
}
\caption{Query conditions and dataset statistics.}
\label{tab:datasets}
\end{table}

\begin{table}[H]
\centering
\scriptsize
\begin{tabular}{lrrrr}
\toprule
\textbf{Statistic} & \textbf{S-Polluted} & \textbf{S-Keys} & \textbf{S-Conv} & \textbf{Real} \\
\midrule
Mean words        & 21.5 & 6.5  & 14.4 & 7.0  \\
Median words      & 19.0 & 6.0  & 14.0 & 5.5  \\
Single word       & 0.0\% & 0.2\% & 0.0\% & 8.8\% \\
2--3 words        & 0.2\% & 10.9\% & 0.6\% & 18.5\% \\
4--7 words        & 4.3\% & 57.9\% & 8.5\% & 42.2\% \\
8--15 words       & 27.8\% & 30.5\% & 51.5\% & 23.9\% \\
16+ words         & 67.7\% & 0.5\% & 39.4\% & 6.6\% \\
Well-formed       & 65.3\% & 38.1\% & 97.5\% & 21.0\% \\
Keyword style     & 39.7\% & 94.8\% & 4.8\%  & 40.4\% \\
Has `?'           & 85.5\% & 10.6\% & 98.7\% & 29.8\% \\
\bottomrule
\end{tabular}
\caption{Extended query characterization for all four conditions ($n=1{,}000$ each).}
\label{tab:query_stats_full}
\end{table}

\section{Oracle Label Distribution}
\label{sec:appendix_oracle_labels}

Table~\ref{tab:oracle_label_dist} shows the distribution of oracle labels across the 2,206 high-contrast queries that form the routing dataset. Each query is assigned to the cheapest workflow that achieves the maximum composite score, subject to a minimum divergence threshold of 1.5 CO points (Section~\ref{sec:routing}). Workflows are grouped into retrieval-only Tier~1 (Semantic, Hybrid, and Semantic-CE) and LLM-augmented Tier~2 (QE-CE and HyDE).

The distribution reveals a consistent pattern. Synthetic conditions skew toward LLM-augmented labels (63--78\%), reflecting the vocabulary gap between verbose synthetic queries and the corpus index. Real user queries are the exception, where the Tier~1 and Tier~2 split narrows to a near-even 47.6\% versus 52.4\%, consistent with the Coverage Illusion finding that retrieval-only workflows serve real users far more often than synthetic benchmarks suggest. HyDE is the dominant oracle label across all four conditions, but its share drops from 74.5\% on Synth-Polluted to 43.3\% on real user queries.

\begin{table}[H]
\centering
\scriptsize
\setlength{\tabcolsep}{4pt}
\resizebox{\columnwidth}{!}{%
\begin{tabular}{lrrrrrrrr}
\toprule
\textbf{Condition} & \textbf{n} & \textbf{Sem.} & \textbf{Hybrid} & \textbf{Sem-CE} & \textbf{QE-CE} & \textbf{HyDE} & \textbf{Tier 1} & \textbf{Tier 2} \\
\midrule
Synth-Polluted  & 542     & 15.5\% &  5.2\% & 0.9\% &  3.9\% & 74.5\% & 21.6\% & 78.4\% \\
Synth-Keywords  & 630     & 15.7\% & 11.7\% & 1.0\% &  4.4\% & 67.1\% & 28.4\% & 71.6\% \\
Synth-Conv.     & 425     & 27.5\% &  7.8\% & 1.4\% &  3.5\% & 59.8\% & 36.7\% & 63.3\% \\
Real User      & 609     & 22.7\% & 21.8\% & 3.1\% &  9.0\% & 43.3\% & 47.6\% & 52.4\% \\
\midrule
All Conditions  & 2{,}206 & 19.9\% & 12.1\% & 1.6\% &  5.4\% & 61.0\% & 33.6\% & 66.4\% \\
\bottomrule
\end{tabular}%
}
\caption{Oracle label distribution across the 2,206 high-contrast routing queries.}
\label{tab:oracle_label_dist}
\end{table}

\section{Human Evaluation Details}
\label{sec:appendix_human_eval}

Three annotators independently scored 100 query--response pairs (5 per workflow $\times$ condition stratum) on faithfulness, answer relevance, and retrieval quality, each on a 1--5 scale. Human scores are averaged across the three annotators before comparison against \texttt{Qwen3-32B}. Table~\ref{tab:human_eval_metrics} reports per-metric agreement statistics; Table~\ref{tab:human_eval_per_wf} reports the aggregate composite overall (CO) score across both judges.

\paragraph{Faithfulness Inter-Annotator Agreement and the Ceiling Effect.}
The near-zero Krippendorff $\alpha$ for faithfulness ($\alpha$=0.01) reflects a score distribution artifact rather than genuine annotator disagreement. Human faithfulness ratings cluster tightly near the ceiling of the scale (mean 4.78 out of 5), leaving almost no variance across items. Krippendorff's $\alpha$ is normalized by expected disagreement, which collapses when the empirical score range is compressed; even a perfectly consistent set of annotators will produce $\alpha \approx 0$ if nearly every item receives the same rating. By contrast, answer relevance and retrieval quality show greater spread (means 4.18 and 4.21) and yield moderate agreement ($\alpha$=0.50 and 0.60, respectively). The averaged human faithfulness signal nonetheless remains informative, as evidenced by its moderate Spearman correlation with the automated judge ($\rho = 0.59, p < 0.001$). Taken together, these results indicate that responses in this deployment are almost uniformly grounded in retrieved sources, making faithfulness a near-constant rather than a discriminative dimension at this operating point.

\begin{table}[H]
\centering
\scriptsize
\resizebox{\columnwidth}{!}{%
\begin{tabular}{lrrrrr}
\toprule
\textbf{Metric} & \textbf{Human} & \textbf{Qwen3-32B} & \textbf{$\Delta$} & \textbf{$\rho$} & \textbf{MAE} \\
\midrule
Faithfulness      & 4.78 & 4.82 & $+$0.04 & 0.59 & 0.18 \\
Answer Relevance  & 4.18 & 4.24 & $+$0.06 & 0.78 & 0.17 \\
Retrieval Quality & 4.21 & 4.35 & $+$0.14 & 0.77 & 0.21 \\
\midrule
\multicolumn{2}{l}{\textit{Inter-annotator (Krippendorff $\alpha$)}} & \multicolumn{4}{r}{Faith.\ 0.01 \quad AnsRel 0.50 \quad RetQ 0.60} \\
\bottomrule
\end{tabular}%
}
\caption{Per-metric agreement between human annotators and \texttt{Qwen3-32B} on 100 sampled query--response pairs. $\Delta$ = Qwen3-32B $-$ Human mean; $\rho$ = Spearman rank correlation between per-query automated scores and mean human scores; MAE is on the 1--5 scale.}
\label{tab:human_eval_metrics}
\end{table}

\begin{table}[H]
\centering
\scriptsize
\resizebox{\columnwidth}{!}{%
\begin{tabular}{lrrrrr}
\toprule
\textbf{Workflow} & \textbf{n} & \textbf{Human CO} & \textbf{Qwen3-32B CO} & \textbf{$\Delta$} & \textbf{$\rho$} \\
\midrule
All         & 100 & 4.48 & 4.53 & $+$0.05 & 0.69 \\
\bottomrule
\end{tabular}%
}
\caption{Aggregate human vs.\ \texttt{Qwen3-32B} CO across all 100 sampled query--response pairs (CO = mean of faithfulness and answer relevance).}
\label{tab:human_eval_per_wf}
\end{table}

\section{LLM Judge Prompt}
\label{sec:appendix_judge_prompt}

The following prompt was used to evaluate all 20,000 query--workflow pairs with \texttt{Qwen3-32B} and \texttt{Qwen3.6-27B}. Template variables \texttt{\{question\}}, \texttt{\{sources\}}, and \texttt{\{answer\}} are filled at inference time.

\begin{lstlisting}
You are evaluating a Danish encyclopedia RAG system (lex.dk).
IMPORTANT RULES:
- You MUST ONLY use the provided sources.
- Do NOT use prior knowledge.
- If a claim is not explicitly supported, treat it as unsupported.
- Be strict: do not give high scores unless clearly justified.
- The question and sources are in Danish. Pay close attention to precise meaning and terminology.
- If NO SOURCES were retrieved, score retrieval_quality=1 with reasoning "No sources retrieved".

QUESTION (in Danish): {question}
RETRIEVED SOURCES:
{sources}
SYSTEM ANSWER:
{answer}
Evaluate THREE aspects:

1. FAITHFULNESS:
Is every claim in the answer supported by the sources?
Evaluation steps:
- Extract the main claims from the answer
- For each claim, check whether it is explicitly supported by the sources
Scoring:
   5: All claims are fully supported by the sources, no hallucination
   4: Mostly supported, only minor unsupported details
   3: Partially supported, some significant unsupported claims
   2: Largely unsupported or introduces incorrect information
   1: Major hallucinations or directly contradicts the sources

2. ANSWER_RELEVANCE:
Does the answer correctly and fully answer the question?
Evaluation steps:
- Identify what the question is asking for
- Check whether the answer addresses ALL key parts
- Penalize missing, vague, or indirect responses
IMPORTANT: If the answer says it cannot find the information, cannot help,
or explains that the sources do not contain the answer -- this scores 1 on
answer_relevance. Acknowledging a limitation is not the same as answering
the question. Do NOT reward a refusal with a high score for being transparent.
Scoring:
   5: Directly and completely answers the question
   4: Mostly answers the question with only minor gaps
   3: Partially answers the question
   2: Addresses the topic but misses the actual question
   1: Off-topic, completely fails to answer, or declines to answer

3. RETRIEVAL_QUALITY:
How relevant are the retrieved sources to the question?
Evaluation steps:
- If NO SOURCES retrieved: ALWAYS score 1 ("No sources retrieved")
- If sources provided:
  * Determine whether the sources are directly about the topic
  * Assess whether they contain the information needed to answer
  * Ignore redundancy; focus only on relevance
Scoring:
   5: All sources are directly about the topic asked -- ideal retrieval
   4: Most sources are relevant, one or two are tangential
   3: Mix of relevant and irrelevant sources
   2: Sources are mostly off-topic, only loosely related to the question
   1: Sources are completely unrelated OR no sources retrieved

Respond with JSON only (no markdown, no explanation outside JSON):
{"faithfulness": {"score": <int>, "reasoning": "<one sentence>"},
 "answer_relevance": {"score": <int>, "reasoning": "<one sentence>"},
 "retrieval_quality": {"score": <int>, "reasoning": "<one sentence>"}}
\end{lstlisting}

\section{Evaluation with Modern Judge Models}
\label{sec:appendix_qwen36}

\begin{table*}[h]
\centering
\small
\begin{tabular}{ll c cccc c cccc}
\toprule
& & & \multicolumn{4}{c}{\textbf{Qwen3-32B}} & & \multicolumn{4}{c}{\textbf{Qwen3.6-27B}} \\
\cmidrule{4-7} \cmidrule{9-12}
\textbf{Workflow} & \textbf{Tier} & \textbf{Cov.} & \textbf{CO} & \textbf{CWA} & \textbf{Faith.} & \textbf{AnsRel} & & \textbf{CO} & \textbf{CWA} & \textbf{Faith.} & \textbf{AnsRel} \\
\midrule
Semantic    & Retr. & 62.6\% & 3.058 & 4.287 & 4.848 & 3.725 & & 2.996 & 4.190 & 4.717 & 3.664 \\
Semantic-CE & Retr. & 62.8\% & 3.086 & 4.321 & 4.865 & 3.777 & & 3.012 & 4.206 & 4.750 & 3.662 \\
Hybrid      & Retr. & 72.2\% & 3.408 & 4.335 & 4.816 & 3.855 & & 3.322 & 4.216 & 4.659 & 3.773 \\
QE-CE       & LLM   & 77.9\% & 3.601 & 4.338 & 4.760 & 3.917 & & 3.493 & 4.205 & 4.597 & 3.813 \\
HyDE        & LLM   & 86.4\% & 3.944 & 4.407 & 4.844 & 3.971 & & 3.836 & 4.285 & 4.692 & 3.878 \\
\bottomrule
\end{tabular}
\caption{Extended baseline performance comparing \texttt{Qwen3-32B} and \texttt{Qwen3.6-27B} on real user queries.}
\label{tab:qwen36_real}
\end{table*}

To check whether our findings hold up under a different judge, we replicated the full evaluation using \texttt{Qwen3.6-27B}. Table~\ref{tab:qwen36_real} places the primary \texttt{Qwen3-32B} results side by side with the \texttt{Qwen3.6-27B} scores on real user queries.

The core picture is unchanged. Retrieval-only Hybrid still reaches 72.2\% coverage with competitive answer quality (CWA=4.216), while HyDE pushes coverage to 86.4\% at the cost of substantial compute. Workflow rankings and the coverage-dominance pattern are consistent across both judges, confirming that the results are not an artifact of one particular scoring model.

\begin{table*}[h]
\centering
\scriptsize
\begin{tabular}{ll rr rr rr rr}
\toprule
& & \multicolumn{2}{c}{\textbf{Synth-Polluted}} & \multicolumn{2}{c}{\textbf{Synth-Keys}} & \multicolumn{2}{c}{\textbf{Synth-Conv}} & \multicolumn{2}{c}{\textbf{Real User}} \\
\cmidrule{3-4} \cmidrule{5-6} \cmidrule{7-8} \cmidrule{9-10}
\textbf{Workflow} & \textbf{n$_{\text{ans}}$} & R\textsuperscript{2}\_cov & R\textsuperscript{2}\_qual & R\textsuperscript{2}\_cov & R\textsuperscript{2}\_qual & R\textsuperscript{2}\_cov & R\textsuperscript{2}\_qual & R\textsuperscript{2}\_cov & R\textsuperscript{2}\_qual \\
\midrule
Semantic    & 573--786 & 0.862 & 0.683 & 0.877 & 0.726 & 0.842 & 0.650 & 0.829 & 0.513 \\
Hybrid      & 623--794 & 0.870 & 0.757 & 0.873 & 0.733 & 0.848 & 0.749 & 0.787 & 0.545 \\
Semantic-CE      & 585--787 & 0.868 & 0.770 & 0.869 & 0.793 & 0.859 & 0.789 & 0.828 & 0.589 \\
QE-CE       & 776--849 & 0.855 & 0.678 & 0.835 & 0.672 & 0.815 & 0.686 & 0.741 & 0.487 \\
HyDE        & 864--954 & 0.772 & 0.528 & 0.668 & 0.545 & 0.774 & 0.553 & 0.685 & 0.504 \\
\midrule
\multicolumn{2}{l}{\textit{Mean R\textsuperscript{2}\_cov}} & \multicolumn{2}{c}{0.845} & \multicolumn{2}{c}{0.824} & \multicolumn{2}{c}{0.828} & \multicolumn{2}{c}{0.774} \\
\bottomrule
\end{tabular}
\caption{Full variance decomposition. R\textsuperscript{2}\_cov is the fraction of CO variance explained by the binary \texttt{has\_sources} flag (retrieved something vs.\ nothing; $n\approx1{,}000$). R\textsuperscript{2}\_qual is the fraction of CO variance explained by the continuous \texttt{retrieval\_quality} score, restricted to answered queries ($n_{\text{ans}}$ shown as range). Declining R\textsuperscript{2}\_cov for HyDE reflects reduced variance in the binary flag as coverage approaches 1.}
\label{tab:variance_full}
\end{table*}

\section{Full Variance Decomposition}
\label{sec:appendix_variance}

Table~\ref{tab:variance_full} reports two statistics for each workflow. First, how much of the spread in CO scores is explained by knowing only whether retrieval succeeded or failed? This fraction is \emph{R\textsuperscript{2}\_cov}, computed as the squared correlation between CO and the binary \texttt{has\_sources} flag. Because CO is fixed at 1.0 whenever \texttt{has\_sources}~$=0$ and is the full quality score otherwise, a high R\textsuperscript{2}\_cov means that the simple success-or-failure outcome accounts for most of the quality variance. Second, among queries that did return documents, how much additional CO variance is explained by \texttt{retrieval\_quality}, the judge's rating of document relevance? This fraction is \emph{R\textsuperscript{2}\_qual}. Throughout the table, R\textsuperscript{2}\_cov $\gg$ R\textsuperscript{2}\_qual. Whether any document was found dominates quality far more than how good those documents were. For real-world RAG, the binary success of locating any relevant source is a far stronger predictor of quality than the incremental precision of the ranking itself.

\section{HyDE Advantage Decomposition and Coverage-Fixed Quality}
\label{sec:appendix_decomp}
\label{sec:appendix_paired}

\paragraph{HyDE Advantage Decomposition.}
Table~\ref{tab:hyde_decomp} breaks down the CO advantage of HyDE over Semantic into a coverage component and an answer quality component, using first-order approximation
\[
\Delta\text{CO} \approx \underbrace{\Delta\text{cov} \times \overline{Q}_{\text{sem}}}_{\text{Via coverage}} + \underbrace{\text{cov}_{\text{sem}} \times \Delta\overline{Q}_{\text{ans}}}_{\text{Via quality}} + \text{residual}
\]
where the terms are defined as:
\begin{itemize}
  \item $\Delta\text{cov}=\text{cov}_{\text{HyDE}}-\text{cov}_{\text{sem}}$, the coverage gain (86.4\%$-$62.6\%$=23.8\%$, Table~\ref{tab:qwen36_real}),
  \item $\overline{Q}_{\text{sem}}$ is Semantic's mean per-answer quality, i.e.\ CWA $= 4.287$ (Table~\ref{tab:qwen36_real}),
  \item $\text{cov}_{\text{sem}}$ is Semantic's coverage rate ($= 62.6\%$),
  \item $\Delta\overline{Q}_{\text{ans}} = \overline{Q}_{\text{HyDE}} - \overline{Q}_{\text{sem}}$ is the gap in mean per-answer quality between HyDE and Semantic ($4.407 - 4.287 = 0.120$), and
  \item \textit{residual} $= \Delta\text{CO} - (\Delta\text{cov} \times \overline{Q}_{\text{sem}} + \text{cov}_{\text{sem}} \times \Delta\overline{Q}_{\text{ans}})$, the gap between the observed $\Delta\text{CO}$ and the sum of the two first-order terms. Numerically, $0.886 - (1.020 + 0.076) = -0.209$. It is negative because the two terms together overestimate the gain, as HyDE's extra coverage comes from harder queries whose actual per-answer quality is below $\overline{Q}_{\text{sem}}$, the value the approximation uses for them.
\end{itemize}

Substituting the real user query values gives a concrete numerical instantiation (Table~\ref{tab:hyde_decomp})
\begin{align*}
\Delta\text{CO} &\approx
  \underbrace{0.238 \times 4.287}_{+1.020}
+ \underbrace{0.626 \times 0.120}_{+0.076}
+ (-0.209) \\
&= +0.886
\end{align*}
The coverage term alone ($+1.020$) exceeds the total observed gap ($+0.886$), while the quality term contributes only $+0.076$. The negative residual ($- 0.209$) accounts for HyDE's harder-query margin. Those marginal queries yield lower per-answer quality than Semantic's average, so the two primary terms together overshoot the actual gain. The practical implication is that HyDE's superiority over Semantic derives almost entirely from retrieving sources for queries that Semantic leaves unanswered, not from generating better answers.

\begin{table}[H]
\centering
\scriptsize
\begin{tabular}{lrrrr}
\toprule
\textbf{Condition} & \textbf{$\Delta$CO} & \textbf{Via cov.} & \textbf{Via qual.} & \textbf{Resid.} \\
\midrule
Synth-Polluted & +1.304 & +1.315 & +0.193 & $-$0.203 \\
Synth-Keys     & +1.599 & +1.712 & +0.160 & $-$0.272 \\
Synth-Conv.    & +0.708 & +0.707 & +0.126 & $-$0.125 \\
Real User     & +0.886 & +1.020 & +0.076 & $-$0.209 \\
\bottomrule
\end{tabular}
\caption{HyDE vs.\ Semantic CO advantage decomposition. ``Via cov.'' and ``Via qual.'' are the two first-order terms; ``Residual'' is the interaction term from harder queries at the margin of HyDE's extra coverage.}
\label{tab:hyde_decomp}
\end{table}

\paragraph{Coverage-Fixed Quality.}
To isolate generation quality from retrieval efficacy, Table~\ref{tab:paired_quality} presents a paired analysis of queries that all workflows answered. Maintaining 100\% coverage, we can assess whether LLM-augmented strategies offer any advantage in answer synthesis. The HyDE vs.\ Semantic answer-relevance gap and its statistical significance are reported in Table~\ref{tab:wilcoxon_summary}. These gaps are small compared to the composite score differences in Table~\ref{tab:baselines}, confirming that per-answer quality is rarely the deciding factor. 

\begin{table}[H]
\centering
\resizebox{\columnwidth}{!}{%
\begin{tabular}{ll rrrr}
\toprule
\textbf{Condition} & \textbf{Workflow} & \textbf{AnsRel} & \textbf{RetQual} & \textbf{Faith.} & \textbf{n} \\
\midrule
\multirow{6}{*}{Real User}
  & Semantic    & 4.086 & 4.558 & 4.846 & 500 \\
  & Hybrid      & 4.206 & 4.614 & 4.826 & 500 \\
  & Semantic-CE      & 4.162 & 4.568 & 4.872 & 500 \\
  & QE-CE       & 4.236 & 4.656 & 4.784 & 500 \\
  & HyDE        & 4.342 & 4.734 & 4.850 & 500 \\
  & \multicolumn{2}{l}{\textit{HyDE$-$Sem $\Delta$AnsRel}} & \multicolumn{2}{r}{$+0.256$ ($p<0.0001$)} & \\
\midrule
\multirow{6}{*}{Synth-Keys}
  & Semantic    & 4.423 & 4.618 & 4.919 & 471 \\
  & Hybrid      & 4.504 & 4.651 & 4.900 & 470 \\
  & Semantic-CE      & 4.430 & 4.589 & 4.951 & 472 \\
  & QE-CE       & 4.594 & 4.754 & 4.928 & 471 \\
  & HyDE        & 4.655 & 4.828 & 4.909 & 472 \\
  & \multicolumn{2}{l}{\textit{HyDE$-$Sem $\Delta$AnsRel}} & \multicolumn{2}{r}{$+0.231$ ($p<0.001$)} & \\
\midrule
\multirow{6}{*}{Synth-Polluted}
  & Semantic    & 4.573 & 4.711 & 4.908 & 543 \\
  & Hybrid      & 4.635 & 4.729 & 4.941 & 543 \\
  & Semantic-CE      & 4.592 & 4.697 & 4.923 & 542 \\
  & QE-CE       & 4.670 & 4.777 & 4.928 & 542 \\
  & HyDE        & 4.810 & 4.884 & 4.943 & 541 \\
  & \multicolumn{2}{l}{\textit{HyDE$-$Sem $\Delta$AnsRel}} & \multicolumn{2}{r}{$+0.235$ ($p<0.001$)} & \\
\midrule
\multirow{6}{*}{Synth-Conv.}
  & Semantic    & 4.709 & 4.821 & 4.925 & 677 \\
  & Hybrid      & 4.732 & 4.827 & 4.931 & 678 \\
  & Semantic-CE      & 4.742 & 4.810 & 4.944 & 679 \\
  & QE-CE       & 4.701 & 4.807 & 4.917 & 678 \\
  & HyDE        & 4.751 & 4.875 & 4.951 & 679 \\
  & \multicolumn{2}{l}{\textit{HyDE$-$Sem $\Delta$AnsRel}} & \multicolumn{2}{r}{$+0.041$ ($p=0.389$)} & \\
\bottomrule
\end{tabular}%
}
\caption{Coverage-fixed quality for queries answered by all five workflows simultaneously ($n$ varies by condition). Coverage is held at 100\% across all workflows, so any differences reflect genuine per-answer quality.}
\label{tab:paired_quality}
\end{table}

\paragraph{Wilcoxon Signed-Rank Test.}
All pairwise comparisons use the Wilcoxon signed-rank test, a non-parametric paired test appropriate for the 1--5 CO scale. For the cascade-vs-HyDE comparison, the test is one-sided, as we test specifically for improvement rather than any difference. Table~\ref{tab:wilcoxon_summary} reports $n$ (query pairs), $\Delta$ (mean per-query score difference), $W$ (the signed-rank statistic), and $p$ (one-sided for cascade comparisons, two-sided otherwise).

\begin{table}[H]
\centering
\scriptsize
\resizebox{\columnwidth}{!}{%
\begin{tabular}{llrrrr}
\toprule
\textbf{Comparison} & \textbf{Condition} & \textbf{n} & \textbf{$\Delta$} & \textbf{W} & \textbf{$p$} \\
\midrule
\multicolumn{6}{l}{\textit{HyDE vs.\ Semantic, answer relevance (paired queries)}} \\
\midrule
HyDE $>$ Sem AnsRel & Real User     & 500     & +0.256 & 2{,}402  & $<$0.001 \\
HyDE $>$ Sem AnsRel & Synth-Keys     & 471     & +0.231 & 1{,}612  & $<$0.001 \\
HyDE $>$ Sem AnsRel & Synth-Polluted & 541     & +0.235 & 951      & $<$0.001 \\
HyDE $>$ Sem AnsRel & Synth-Conv.    & 677     & +0.041 & 2{,}098  & $0.389$  \\
\midrule
\multicolumn{6}{l}{\textit{Cascade vs.\ Always-HyDE, composite overall}} \\
\midrule
\textbf{Cascade $>$ HyDE CO} & \textbf{Real User}     & \textbf{1{,}000} & \textbf{+0.140} & \textbf{38{,}235} & \textbf{$<$0.001} \\
\bottomrule
\end{tabular}%
}
\caption{Wilcoxon signed-rank test results for all pairwise comparisons reported in this paper.}
\label{tab:wilcoxon_summary}
\end{table}

\section{R\textsuperscript{2} Sensitivity Analysis}
\label{sec:appendix_sensitivity}

To address the potential concern that the high $R^2$ values are a mathematical artifact of the 1.0 scoring floor, Table~\ref{tab:r2_sensitivity} provides a sensitivity analysis using increasingly stringent ``success'' thresholds. We treat a query as covered only if the retrieved sources meet a minimum quality grade ($\geq$2, 3, or 4). Coverage remains a strong predictor across thresholds, indicating that the dominance of the binary success signal is intrinsic to the RAG system rather than a consequence of the scoring scale.

\begin{table}[H]
\centering
\scriptsize
\begin{tabular}{llrrrr}
\toprule
\textbf{Workflow} & \textbf{Threshold} & \textbf{Cov.\ Rate} & \textbf{R\textsuperscript{2}\_cov} \\
\midrule
\multirow{4}{*}{Semantic}
  & Binary (\texttt{has\_sources}) & 62.6\% & 0.829 \\
  & Quality $\geq$ 2               & 58.5\% & 0.804 \\
  & Quality $\geq$ 3               & 55.4\% & 0.794 \\
  & Quality $\geq$ 4               & 48.5\% & 0.781 \\
\midrule
\multirow{4}{*}{Hybrid}
  & Binary (\texttt{has\_sources}) & 72.2\% & 0.787 \\
  & Quality $\geq$ 2               & 67.3\% & 0.767 \\
  & Quality $\geq$ 3               & 63.9\% & 0.767 \\
  & Quality $\geq$ 4               & 56.9\% & 0.766 \\
\midrule
\multirow{4}{*}{QE-CE}
  & Binary (\texttt{has\_sources}) & 77.9\% & 0.741 \\
  & Quality $\geq$ 2               & 74.1\% & 0.727 \\
  & Quality $\geq$ 3               & 70.4\% & 0.734 \\
  & Quality $\geq$ 4               & 63.3\% & 0.743 \\
\midrule
\multirow{4}{*}{HyDE}
  & Binary (\texttt{has\_sources}) & 86.4\% & 0.685 \\
  & Quality $\geq$ 2               & 84.2\% & 0.669 \\
  & Quality $\geq$ 3               & 80.5\% & 0.668 \\
  & Quality $\geq$ 4               & 72.5\% & 0.699 \\
\bottomrule
\end{tabular}
\caption{R\textsuperscript{2} sensitivity analysis (real user queries). R\textsuperscript{2} remains stable as the coverage threshold tightens.}
\label{tab:r2_sensitivity}
\end{table}

\section{Regression Router Details}
\label{sec:appendix_tables}

\begin{table}[H]
\centering
\small
\resizebox{\columnwidth}{!}{%
\begin{tabular}{rrrrrr}
\toprule
$\delta$ & \textbf{Router CO} & $\Delta$ & \textbf{Savings} & \textbf{Oracle Gap}  \\
\midrule
0.00 & 4.200 & $-$0.015 & 13.9\% & $-$3.4\% \\
0.50 & 4.230 & +0.015   &  6.1\% &  2.9\%   \\
\textbf{0.75} & \textbf{4.233} & \textbf{+0.018} & \textbf{4.1\%} & \textbf{3.7\%} \\
1.00 & 4.225 & +0.011   &  2.4\% &  2.2\%   \\
2.50 & 4.215 & +0.000   &  0.0\% &  0.0\%   \\
\midrule
Oracle & 4.720 & +0.505  & 39.0\% & 100.0\%  \\
\bottomrule
\end{tabular}%
}
\caption{Quality--efficiency trade-off for the Ridge regression router under 5-fold cross-validation. $\delta$ is the confidence margin required before substituting a cheaper workflow for HyDE; $\Delta$ is composite overall relative to Always-HyDE; Savings is the fraction of queries redirected away from HyDE; Oracle Gap is the percentage of the HyDE--oracle gap recovered.}
\label{tab:regression_pareto}
\end{table}

\section{Cross-Condition Performance}
\label{sec:appendix_cross_condition}

Table~\ref{tab:cross_condition} evaluates the cascade against Always-HyDE on the full 1,000-query set for each condition. Table~\ref{tab:cross_condition_routing} repeats the comparison restricted to the 2,206-query high-contrast routing subset from Section~\ref{sec:routing}, in which each query's best and worst workflows differ by at least 1.5 CO points. Comparing the two tables reveals the robustness of the cascade advantage on real user queries. The gap increases from $+0.140$ (full set) to $+0.241$ (routing subset) because the routing subset retains only queries in which the workflow choice materially changes the outcome.

\begin{table}[H]
\centering
\scriptsize
\begin{tabular}{lrrrrl}
\toprule
\textbf{Condition} & \textbf{Cascade} & \textbf{HyDE} & \textbf{$\Delta$} & \textbf{Saved} & \textbf{Verdict} \\
\midrule
Real User     & 4.084 & 3.944 & +0.140  & 31.8\% & Cascade \\
Synth-Polluted & 4.559 & 4.607 & $-$0.048 & 29.8\% & HyDE \\
Synth-Keywords & 4.471 & 4.610 & $-$0.139 & 29.1\% & HyDE \\
Synth-Conv.    & 4.620 & 4.610 & +0.010  & 41.9\% & Cascade \\
\bottomrule
\end{tabular}
\caption{Cascade vs.\ Always-HyDE across all four conditions, each evaluated on the full 1,000-query set.}
\label{tab:cross_condition}
\end{table}

\begin{table}[H]
\centering
\scriptsize
\begin{tabular}{lrrrrl}
\toprule
\textbf{Condition} & \textbf{n} & \textbf{Cascade} & \textbf{HyDE} & \textbf{$\Delta$} & \textbf{Verdict} \\
\midrule
Real User     & 609  & 3.936 & 3.695 & $+$0.241  & Cascade \\
Synth-Polluted & 542  & 4.315 & 4.417 & $-$0.102  & HyDE \\
Synth-Keywords & 630  & 4.314 & 4.534 & $-$0.221  & HyDE \\
Synth-Conv.    & 425  & 4.249 & 4.227 & $+$0.022  & Cascade \\
\bottomrule
\end{tabular}
\caption{Cascade vs.\ Always-HyDE restricted to the 2,206 high-contrast queries (Section~\ref{sec:routing}), where best and worst workflow diverge by at least 1.5 CO points. The cascade advantage on real user queries grows from $+0.140$ (full set) to $+0.241$ (routing subset).}
\label{tab:cross_condition_routing}
\end{table}

\section{Head-to-Head Quality Analysis of Hybrid and HyDE on Shared Queries}
\label{sec:appendix_hybrid_vs_hyde}

Table~\ref{tab:hybrid_vs_hyde} shows the per-query outcome distribution for the 722 real user queries where Hybrid retrieved at least one source. For each query, Hybrid's CO score is compared directly with HyDE's.

\begin{table}[H]
\centering
\scriptsize
\begin{tabular}{lrr}
\toprule
\textbf{Outcome} & \textbf{\% of queries} & \textbf{Mean $|\Delta|$ CO} \\
\midrule
Identical scores (tie) & 59.6\% & 0.000 \\
Hybrid beats HyDE      & 21.2\% & 2.010 \\
HyDE beats Hybrid      & 19.3\% & 1.324 \\
\midrule
\multicolumn{2}{l}{\textit{Net aggregate score flow (Hybrid)}} & +123.5 pts \\
\bottomrule
\end{tabular}
\caption{Pairwise quality comparison on 722 real user queries where Hybrid found sources. When Hybrid beats HyDE, it does so by a larger margin than when HyDE wins, producing a net Hybrid advantage of 123.5 aggregate CO points. LLM-augmented retrieval hurts more than it helps when keyword matching has already located the relevant article.}
\label{tab:hybrid_vs_hyde}
\end{table}

\section{AI Assistance in Research and Writing}
\label{sec:ai_assistants}
We used AI tools to assist with code generation, debugging, data analysis, spell-checking, and grammatical editing. Specifically, we used Anthropic’s \textit{Claude Sonnet 4.6} and \textit{Claude Opus 4.7}.

\section{Licenses}
\label{sec:licenses}
All models are used consistent with their intended purpose. Open models carry the following licenses: \texttt{multilingual-e5-large} (MIT), \texttt{Qwen3-32B} and \texttt{Qwen3.6-27B} (Apache~2.0), \texttt{all-MiniLM-L6-v2} and \texttt{paraphrase-multilingual-MiniLM-L12-v2} (Apache~2.0), and \texttt{gemma-4-26B-A4B-it} (Gemma Terms of Use). GPT-5.2 and GPT-3.5 Turbo were accessed via the OpenAI API within its permitted research use cases. Production query logs were accessed under institutional agreement with the Danish National Encyclopedia, used solely for non-commercial research, and consist of bare search strings with no user identifiers or personally identifiable information. The encyclopedia context structurally limits offensive content risk, as queries are factual topic lookups against a curated reference corpus. The evaluation dataset and trained routing models are not publicly released.

\end{document}